%% file: acl_latex.tex
\pdfoutput=1
\documentclass[11pt]{article}

\usepackage{acl}

\usepackage{times}
\usepackage{latexsym}

\usepackage[T1]{fontenc}
\usepackage[utf8]{inputenc}

\usepackage{microtype}

\usepackage{amsmath}
\usepackage{amsfonts}
\usepackage{bbm}

\newcommand*{\E}{\mathbb{E}}
\newcommand{\given}{\,\vert\,}
\newcommand{\param}{\,;\,}

\newcommand{\ashift}{\textsc{shift}}
\newcommand{\anode}{\textsc{node}}
\newcommand{\acopy}{\textsc{copy}}
\newcommand{\ala}{\textsc{la}}
\newcommand{\ara}{\textsc{ra}}
\newcommand{\aeos}{\textsc{end}}
\newcommand{\oracle}{O(l, w, g)}
\DeclareMathOperator*{\deff}{def}
\newcommand\myeq{\mathrel{\overset{\makebox[0pt]{\mbox{\normalfont\tiny $\deff$}}}{=}}}

\usepackage{pifont}
\usepackage{newpxmath}

\DeclareMathOperator*{\KL}{KL}

\DeclareMathOperator*{\argmax}{arg\,max}

\usepackage{booktabs}
\usepackage{fancyvrb}
\usepackage{xcolor}
\usepackage{tikz}
\usetikzlibrary{positioning}
\usetikzlibrary{arrows.meta,arrows}

\title{Inducing and Using Alignments for Transition-based AMR Parsing}

\author{Andrew Drozdov$^\text{\tiny\textdagger}$ \ \ \
    Jiawei Zhou$^\text{\tiny\textdaggerdbl}$ \ \ \
    Radu Florian$^{\scriptscriptstyle\Diamondblack}$ \ \ \
    Andrew McCallum$^\text{\tiny\textdagger}$ \\
    {\bf Tahira Naseem$^{\scriptscriptstyle\Diamondblack}$ \ \ \
    Yoon Kim$^{\scriptscriptstyle\Diamond}$ \ \ \
    Ramon Fernandez Astudillo$^{\scriptscriptstyle\Diamondblack}$}\\
  \AND\\[-4ex]
$^\text{\tiny\textdagger}$UMass Amherst CICS \hspace{1cm} 
$^\text{\tiny\textdaggerdbl}$Harvard University \hspace{1cm}
$^{\scriptscriptstyle\Diamond}$MIT CSAIL \hspace{1cm}
$^{\scriptscriptstyle\Diamondblack}$IBM Research   \\
\AND\\[-4ex]
$^\text{\tiny\textdagger}$\texttt{\small{\{adrozdov,mccallum\}@cs.umass.edu}} \ \ 
$^\text{\tiny\textdaggerdbl}$\texttt{\small{jzhou02@g.harvard.edu}}
\\
$^{\scriptscriptstyle\Diamond}$\texttt{\small{yoonkim@mit.edu}} \ \
$^{\scriptscriptstyle\Diamondblack}$\texttt{\small{\{raduf,tnaseem\}@us.ibm.com}} \ \
$^{\scriptscriptstyle\Diamondblack}$\texttt{\small{ramon.astudillo@ibm.com}}
}

\begin{document}
\maketitle

\begin{abstract}

Transition-based parsers for Abstract Meaning Representation (AMR) rely on node-to-word alignments. These alignments are learned separately from parser training and require a complex pipeline of rule-based components, pre-processing, and post-processing to satisfy domain-specific constraints. Parsers also train on a point-estimate of the alignment pipeline, neglecting the uncertainty due to the inherent ambiguity of alignment. In this work we explore two avenues for overcoming these limitations. First, we propose a neural aligner for AMR that learns node-to-word alignments without relying on complex pipelines. We subsequently explore a tighter integration of aligner and parser training by considering a distribution over oracle action sequences arising from aligner uncertainty. Empirical results show this approach leads to more accurate alignments and generalization better from the AMR2.0 to AMR3.0 corpora. We attain a new state-of-the art for gold-only trained models, matching silver-trained performance without the need for beam search on AMR3.0. 

\end{abstract}

\input{intro}
\input{amrparsing}

\input{neuralaligner}

\input{oraclesampling}

\input{experiments}

\input{relatedwork}

\section{Conclusion}

In this work we propose a general-purpose neural AMR aligner, which does not use a complex alignment pipeline and generalizes well across domains. The neural parameterization allows the aligner to fully condition on the source and target context and easily incorporates pretrained embeddings, leading to improved performance. Simply using our aligner to produce training data for a state-of-the-art transition-based parser leads to improved results.

We additionally propose a learning procedure using posterior regularization  and importance sampling that involves sampling different action sequences during training. These incorporate alignment uncertainty and further improve parser performance.
Our results on gold-only AMR training (i.e., without silver data) show that parsers learned this way match the performance of the prior state-of-the-art parsers without requiring beam search at test time.

\section*{Acknowledgements}

We are grateful to our colleagues for their  help and advice, and to the anonymous reviewers at ACL Rolling Review for their feedback on drafts of this work. We also thank Nathan Schneider and Austin Blodgett for sharing the gold alignment data. AD and AM were supported in part by the Center for Intelligent Information Retrieval and the Center for Data Science; in part by the IBM Research AI through the AI Horizons Network; in part by the Chan Zuckerberg Initiative under the project Scientific Knowledge Base Construction; in part by the National Science Foundation (NSF) grant numbers  IIS-1922090, IIS-1955567, and IIS-1763618; in part by the Defense Advanced Research Projects Agency (DARPA) via Contract No. FA8750-17-C-0106 under Subaward No. 89341790 from the University of Southern California; and in part by the Office of Naval Research (ONR) via Contract No. N660011924032 under Subaward No. 123875727 from the University of Southern California. YK was supported in part by a MIT-IBM Watson AI grant. Any opinions, findings and conclusions or recommendations expressed in this material are those of the authors and do not necessarily reflect those of the sponsor.

\section*{Ethical Considerations}

We do not foresee specific risks associated with our exploration of alignment for AMR parsing. That being said, there are nuances to our results that future researchers may take into considerations. For instance, our experiments are only for English data in the genres covered by AMR2.0 and AMR3.0. It is not clear how our results translate to other domains (e.g. biomedical text) or other languages. Nonetheless, we are hopeful that are methods would transfer favorably because they are intentionally designed to be easy to use and general purpose.

% Entries for the entire Anthology, followed by custom entries
\bibliographystyle{acl_natbib}
\bibliography{custom.bib}

\clearpage

\appendix

\section{StructBART Implementation Details}
\label{sec:app_structbart}

\noindent Inference in a transition-based parser corresponds to the usual decoding of with a sequence to sequence model

\begin{align*}
\hat{a} = \arg\max_{a} \{ p(a \mid w)\}
\end{align*}

to obtain the graph $\hat{g} = M(\hat{a}, w)$. The model $p(a \mid w)$ is nowadays parametrized with a neural network. The state machine $M(a, w)$ is defined by 

\begin{itemize}
\item the position of a cursor $k \in [|w|] = 1 \cdots |w|$, moving left to right over $w$
\item a stack of nodes of $g$ that is initially empty 
\item the partial graph of $g$
\end{itemize}

Furthermore, each action is decoupled into normal and pointer actions $a_t = (b_t, r_t)$ with

\begin{align*}
&p(a_t \mid a_{<t}, w; \theta) = \\
&\qquad p(r_t \mid b_t, a_{<t}, w; \theta) \, p(b_t \mid a_{<t}, w; \theta)
\end{align*}

\noindent where

\begin{align*}
&p(b_t \mid a_{<t}, w; \theta) =  \\
&\qquad \mathrm{softmax}(\mathrm{NN}_{\theta}(a_{<t}, w) + m(a_{<t}, w))_{b_t}
\end{align*}

where $\mathrm{NN}_{\theta}(a_{<t}, w) \in \mathbb{R}^{|V_b|}$ are the logits coming from a neural network model and $m(a_{<t}, w) \in \{-\infty, 0\}^{V_b}$ is a mask forbidding invalid state machine actions e.g. shifting at sentence end. This mask is deterministically computed given the machine state. $V_b$ is the vocabulary of normal actions (all actions minus pointer information). The pointer network is given by

\begin{align*}
&p(r_t \mid b_t, a_{<t}, w; \theta) = \\ 
&\qquad \mathrm{softmax}(\mathrm{DSA}_{\theta}(a_{<t}, w) + m2(a_{<t}, w))_{b_t})
\end{align*}

for $b_t$ executing arc $\textrm{LA/RA}$ actions and $1$ otherwise. The network $\mathrm{DSA}_{\theta}(a_{<t}, w) \in \mathbb{R}^{|a_{<t}|}$ is the decoder's self-attention encoding of action history $a_{<t}$ (last layer). The mask $m2$ prevents pointing to any action that is not a node generating action, or has been reduced.

\section{Parser Training Details}
\label{sec:app_parse_train}

Training with argmax alignments takes approximately 12 hours on a single GPU (2080ti), and longer when sampling alignments proportional to the number of samples.  When training with samples, we use gradient accumulation to avoid out of memory problems. We choose accumulation steps roughly proportional to number of samples. Otherwise, training hyperparameters exactly match those from \citet{zhou-etal-2021-structure}.

\section{Aligner Implementation Details}
\label{sec:app_align_implement}

\paragraph{Comment on sequence length.} The alignment model in \citet{wu-etal-2018-hard} demonstrates strong results for character-level translation, which involves translating a single word from one language to another character-by-character. They state the following in reference to their alignment model: \textit{the exact marginalization scheme is practically unworkable for machine translation}. As a point of reference, we looked at the inflection dataset --- in any of the training splits across the 51 languages, 99\% of the sequences are less than 29 tokens long, 90\% are less than 23 tokens, and 50\% are less than 15. In contrast, sentence lengths for AMR3.0 are often considerably longer --- the 99/90/50 percentiles for sentence token lengths are 62/35/15 and the AMR token lengths are 45/26/11. Nonetheless, we found their results encouraging and our implementation of the alignment model to still be fast enough despite our using longer sequences.\footnote{We use number of AMR nodes to represent token length, since this is what bounds the computation. }

\section{Aligner Training Details}
\label{sec:app_align_train}

We use single layer LSTMs with size 200 hidden dimension, dropout 0.1, learning rate 0.0001, and train with the Adam Optimizer. We use batch size 32 and accumulate gradient over 4 steps (for an effective batch size of 128). For 200 epochs, training takes approximately 1-day on a single GPU (2080ti). We train a new aligner for each parsing experiment. In Table \ref{tab:amr_alignment_results} we report alignment results from our highest Smatch parsing experiment.

\section{Alternative text for Figure \ref{fig:amr_input_example}}

Shown are the baseline's point estimate alignment and our aligner's alignment posterior. There are instances of ambiguity where our alignment distribution is preferred to the point estimate.

\end{document}

%% file: intro.tex
\section{Introduction}

Abstract Meaning Representation (AMR) was introduced as an effort to unify various semantic tasks (entity-typing, co-reference, relation extraction, and so on; \citealp{banarescu-etal-2013-abstract}). Of existing  approaches for AMR parsing, transition-based parsing is particularly notable because it is high performing but still relies on node-to-word alignments as a core pre-processing step \cite[][inter alia]{ballesteros-al-onaizan-2017-amr,liu-etal-2018-amr,naseem-etal-2019-rewarding,emnlp2020stacktransformer,zhou-etal-2021-amr,zhou-etal-2021-structure}. These alignments are not in the training data and must be learned separately via a complex pipeline of rule-based systems, pre-processing (e.g., lemmatization), and post-processing to satisfy domain-specific constraints. Such pipelines can fail to generalize well, propagating errors into training that reduce AMR performance in new domains (e.g., AMR3.0).
This work studies how we can probabilistically induce and use alignments for transition-based AMR parsing in a domain-agnostic manner, ultimately replacing the existing heuristics-based pipeline.

To \emph{induce} alignments, we propose a neural aligner which uses hard attention within a sequence-to-sequence model to learn latent alignments \cite{wu-etal-2018-hard,deng2018,shankar2018}. While straightforward, this neural parameterization makes it possible to easily incorporate pretrained features such as character-aware word embeddings from ELMo \cite{peters-etal-2018-deep} and also relax some of the strong independence assumptions in classic count-based aligners such as IBM Model 1 \cite{brown1993mathematics}. We find that the neural aligner meaningfully improves upon various baselines, including the existing domain-specific approach.

To \emph{use} the neural aligner's posterior distribution over alignments, we explore several methods. Our first approach simply uses the MAP alignment from the neural aligner to obtain a single oracle action sequence, which is used to train the AMR parser. However, this one-best alignment fails to take into account the inherent uncertainty associated with posterior alignments. Our second approach addresses this via posterior regularization to push the AMR parser's (intractable) posterior alignment distribution to be close to the neural aligner's (tractable) posterior distribution. We show that optimizing this posterior regularized objective results in a simple training scheme wherein the AMR parser is trained on oracle actions derived samples from the neural aligner's posterior distribution. Our final approach uses the neural aligner not as a regularizer but as an importance sampling distribution, which can be used to better approximate samples from the AMR parser's posterior alignment distribution, and thus better approximate the otherwise intractable log marginal likelihood.

In summary, we make the following empirical and methodological contributions:

\begin{itemize}
\item We show that our approach can simplify the existing pipeline and learn state-of-the-art AMR parsers that perform well on both AMR2.0 and AMR3.0. Unlike other approaches, AMR parsers learned this way do not require beam search and hence are more efficient at test time.
\item We explore different methods for inducing and using alignments. We show that a neural parameterization of the aligner is crucial for learning good alignments, and that using the neural aligner to regularize the AMR parser's posterior is an effective strategy for transferring strong inductive biases from the (constrained) aligner to the (overly flexible) parser.
\end{itemize}

%% file: amrparsing.tex
\vspace{-1mm}
\section{Background: Transition-based Parsing}
\label{amrparsing}
\vspace{-1mm}
\subsection{General Approach for AMR}

A standard and effective way to train AMR parsers is with sequence-to-sequence learning where the input sequence is the sentence $w$ and the output sequence is the graph $g$ decomposed into an action sequence $a$ via an oracle. The combination of words and actions is provided to a parameter-less state machine $M$ that produces the graph $g := M(w, a)$. The state machine can perform the \textit{oracle} inverse operation $O$ when also provided alignments $l$, mapping a graph to a deterministic sequence of oracle actions $a := \oracle$.\footnote{While current state-of-the-art oracles do make use of alignments, some oracles do not make explicit use of alignments to derive action sequences, for example by generating the nodes in the AMR graph in depth-first order from the root and breaking ties according to the order nodes appear in the data file.}  During training the model learns to map $w \rightarrow a$ (these pairs are given by the oracle $O$), and $M$ is used to construct graphs ($a \rightarrow g$) for evaluation. 

\subsection{StructBART\label{actionpointingoracle}}

In this paper we use the oracle and state machine from StructBART \cite{zhou-etal-2021-structure}, which is a simplified version of \citet{zhou-etal-2021-amr}. They rely on rules that determine which actions are valid (e.g. the first action can not be to generate an edge). The actions are the output space the parser predicts and when read from left-to-right are used to construct an AMR graph. In this case, the actions incorporate alignments.

\paragraph{Rules} The following rules define the valid actions at each time step:
\begin{itemize}
    \item Maintain a cursor that reads the sentence left-to-right, only progressing for $\ashift$ action.
    \item At each cursor position, generate any nodes aligned to the cursor's word. (This is where node-word alignments are needed).
    \item Immediately after generating a node, also generate any valid incoming or outgoing arcs.
\end{itemize}

\vspace{-1mm}
\paragraph{Actions} At each time step perform one of the following actions to update the state or graph:

\begin{itemize}
    \item $\ashift$: Increment the cursor position.
    \item $\anode(y_v)$: Generate node with label $y_v$.
    \item $\acopy$: Generate node by copying word under the cursor as the label.
    \item $\ala(y_e, n)$, $\ara(y_e, n)$: Generate an edge with label $y_e$ from the most recently generated node to the previously generated node $n$. $\ala$ and $\ara$ (short for left-arc and right-arc) indicate the edge direction as outgoing/incoming. We use $y_e$ to differentiate edge labels from node labels $y_v$.
    \item $\aeos$: A special action indicating that the full graph has been generated.
\end{itemize}

\vspace{-1mm}
\paragraph{Learning} For parsing, StructBART fine-tunes BART \cite{lewis-etal-2020-bart} with the following modifications: a) it converts one attention head from the BART decoder into a \textit{pointer network} for predicting $n$ in the $\ala$/$\ara$ actions, b) logits for actions are masked to guarantee graph well-formedness,  and c) alignment is used to mask two cross-attention heads of the BART decoder,\footnote{Alignment is represented in the action sequence through the $\ashift$ action and cursor position.} thereby integrating structural alignment directly in the model. 

StructBART is trained to optimize the maximum likelihood of action sequences given sentence and alignment. More formally, for a single example $$(w, g, l) \sim \mathcal{D}, \:\:\:a := O(l, w, g),$$
\noindent the log-likelihood of the actions (and hence the graph) is given by,
\begin{align*}
\log p(a \mid w\param \theta) = \sum_{t=1}^T \log p(a_t \mid a_{<t}, w\param \theta)
\label{crossentropy}
\end{align*}

\noindent for a model with parameters $\theta$. Probabilities of actions that create arcs are decomposed into independent label and pointer distributions
\begin{align*}
&p(a_t \mid a_{<t}, w\param \theta) = \\
&\qquad p(y_e \mid a_{<t}, w\param \theta) \, p(n \mid a_{<t}, w\param \theta)
\end{align*} 
where $p(y_e \mid a_{<t}, w\param \theta)$ is computed with the normal output vocabulary distribution of BART and $p(n \mid a_{<t}, w\param \theta)$ with one attention head of the decoder. See \citet{zhou-etal-2021-structure} for more details. 
\vspace{-1mm}
\paragraph{Alignment (SB-Align)}
For training, StructBART depends on node-to-word AMR alignments $l$ to specify the oracle actions. In previous work, the alignments have been computed by a pipeline of components that we call SB-Align.

We introduce our neural approach in the next section, but first we cover the main steps in SB-Align:  (1) produce initial alignments using Symmetrized Expectation Maximization \cite{pourdamghani-etal-2014-aligning}; (2) attempt to align additional nodes by inheriting child node alignments; (3) continue to refine alignments using JAMR \cite{flanigan-etal-2014-discriminative}, which involves constraint optimization using a set of linguistically motivated rules.

The StructBART action space requires that all nodes are aligned, yet after running SB-Align some nodes are not. This is solved by first ``force aligning'' unaligned nodes to unaligned tokens, then propagating alignments from child-to-parent nodes and vice versa until $100\%$ of nodes are aligned to text spans. Finally, node-to-span alignments are converted into node-to-token alignments for model training (e.g. by deterministically aligning to the first node of an entity). Specifics are described in StructBART and preceding work  \cite{zhou-etal-2021-structure,zhou-etal-2021-amr,emnlp2020stacktransformer}.

%% file: neuralaligner.tex
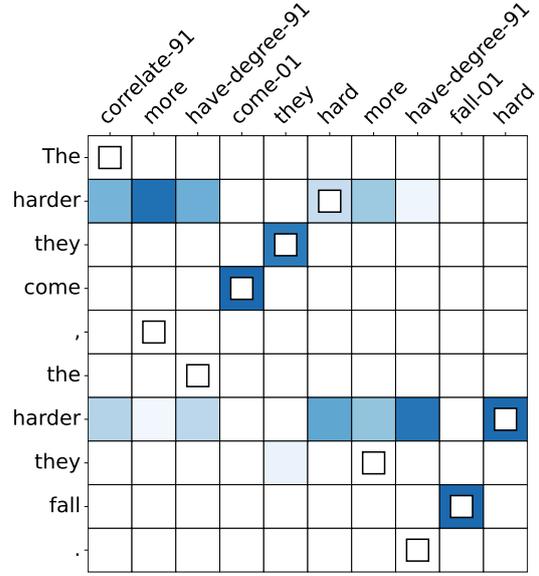
\begin{figure*}
\begin{minipage}{0.49\linewidth}
\textbf{Input sentence}
\vspace{-2mm}

{%fontsize
\fontsize{8pt}{10pt}\selectfont
\begin{Verbatim}[commandchars=\\\{\}]
\textbf{The harder they come , the harder they fall .}
\end{Verbatim}
}

\textbf{AMR Graph}
\vspace{1.5mm}

\resizebox{\linewidth}{!}{%
{
\begin{tikzpicture}[]

% Nodes
\node[draw,style={draw=white,fill=white}] at (0, 0) (c)  {\bf\large correlate-91};
\node[draw,style={draw=white,fill=white}] at (4, 0) (m)  {\bf\large more};
\node[draw,style={draw=white,fill=white}] at (8, 0) (h3)  {\bf\large have-degree-91};
\node[draw,style={draw=white,fill=white}] at (12, 0) (h)  {\bf\large hard};

\node[draw,style={draw=white,fill=white}] at (2, -2) (m2)  {\bf\large more};
\node[draw,style={draw=white,fill=white}] at (6, -2) (h2)  {\bf\large hard};
\node[draw,style={draw=white,fill=white}] at (10, -2) (c2)  {\bf\large come-01};

\node[draw,style={draw=white,fill=white}] at (4, -4) (m)  {\bf\large have-degree-91};
\node[draw,style={draw=white,fill=white}] at (8, -4) (h3)  {\bf\large fall-01};
\node[draw,style={draw=white,fill=white}] at (12, -4) (t)  {\bf\large they};

% Boxes
\draw[rounded corners, line width=0.5mm] (-1.25, -0.5) rectangle (1.25, 0.5) {};
\draw[rounded corners, line width=0.5mm] (3.25, -0.5) rectangle (4.75, 0.5) {};
\draw[rounded corners, line width=0.5mm] (6.5, -0.5) rectangle (9.5, 0.5) {};
\draw[rounded corners, line width=0.5mm] (11.25, -0.5) rectangle (12.75, 0.5) {};

\draw[rounded corners, line width=0.5mm] (1.25, -2.5) rectangle (2.75, -1.5) {};
\draw[rounded corners, line width=0.5mm] (5.25, -2.5) rectangle (6.75, -1.5) {};
\draw[rounded corners, line width=0.5mm] (9, -2.5) rectangle (11, -1.5) {};

\draw[rounded corners, line width=0.5mm] (2.5, -4.5) rectangle (5.5, -3.5) {};
\draw[rounded corners, line width=0.5mm] (7.125, -4.5) rectangle (8.875, -3.5) {};
\draw[rounded corners, line width=0.5mm] (11.25, -4.5) rectangle (12.75, -3.5) {};

% Edges
\draw[-{Latex[length=2mm, width=2mm]}, line width=0.5mm] (1.25, 0) -- (3.25, 0);
\draw[-{Latex[length=2mm, width=2mm]}, line width=0.5mm] (4.75, 0) -- (6.5, 0);
\draw[-{Latex[length=2mm, width=2mm]}, line width=0.5mm] (9.5, 0) -- (11.25, 0);

\draw[-{Latex[length=2mm, width=2mm]}, line width=0.5mm] (0.25, -0.5) -- (1.75, -1.5);
\draw[-{Latex[length=2mm, width=2mm]}, line width=0.5mm] (8.25, -0.5) -- (9.75, -1.5);
\draw[-{Latex[length=2mm, width=2mm]}, line width=0.5mm] (11.75, -0.5) -- (10.25, -1.5);

\draw[-{Latex[length=2mm, width=2mm]}, line width=0.5mm] (2.25, -2.5) -- (3.75, -3.5);
\draw[-{Latex[length=2mm, width=2mm]}, line width=0.5mm] (4.25, -3.5) -- (5.75, -2.5); % 
\draw[-{Latex[length=2mm, width=2mm]}, line width=0.5mm] (6.25, -2.5) -- (7.75, -3.5); 
\draw[-{Latex[length=2mm, width=2mm]}, line width=0.5mm] (10.25, -2.5) -- (11.75, -3.5);

\draw[-{Latex[length=2mm, width=2mm]}, line width=0.5mm] (5.5, -4) -- (7.125, -4);
\draw[-{Latex[length=2mm, width=2mm]}, line width=0.5mm] (8.875, -4) -- (11.25, -4);

% Edge Labels
\node[draw,style={draw=none,fill=none}] at (2.25, 0.25) (a)  {\bf arg1};
\node[draw,style={draw=none,fill=none}] at (5.6, 0.25) (a)  {\bf arg3-of};
\node[draw,style={draw=none,fill=none}] at (10.4, 0.25) (a)  {\bf arg1};

\node[draw,style={draw=none,fill=none},anchor=east] at (0.75,  -1) (a)  {\bf arg2};
\node[draw,style={draw=none,fill=none},anchor=east] at (8.75,  -1) (a)  {\bf arg1};
\node[draw,style={draw=none,fill=none},anchor=west] at (11.25, -1) (a)  {\bf manner-of};

\node[draw,style={draw=none,fill=none},anchor=east] at (2.75,  -3) (a)  {\bf arg3-of};
\node[draw,style={draw=none,fill=none},anchor=east] at (4.75,     -3) (a)  {\bf arg2};
\node[draw,style={draw=none,fill=none},anchor=west] at (7.25,  -3) (a)  {\bf manner-of};
\node[draw,style={draw=none,fill=none},anchor=west] at (11.25, -3) (a)  {\bf arg1};

\node[draw,style={draw=none,fill=none}] at (6.25, -3.75) (a)  {\bf arg1};
\node[draw,style={draw=none,fill=none}] at (10, -3.75) (a)  {\bf arg1};

\end{tikzpicture}
}}%resize
\vspace{2mm} \\
\textbf{Linearized AMR Tree}
\vspace{-2mm}

{%fontsize
\fontsize{8pt}{10pt}\selectfont
\begin{Verbatim}[commandchars=\\\{\}]
( \textbf{correlate-91} :ARG1 ( \textbf{more} :ARG3-of 
    ( \textbf{have-degree-91} :ARG1 ( \textbf{come-01} :ARG1 
    ( \textbf{they} ) :manner ( \textbf{hard} ) ) ) ) :ARG2
    ( \textbf{more} :ARG3-of ( \textbf{have-degree-91} :ARG1 
    ( \textbf{fall-01} :manner ( \textbf{hard} ) ) ) ) )
\end{Verbatim}
}

\end{minipage}
\hfill
\begin{minipage}{0.49\linewidth}

\includegraphics[width=\linewidth,trim={0 -0.1cm 0 0},clip]{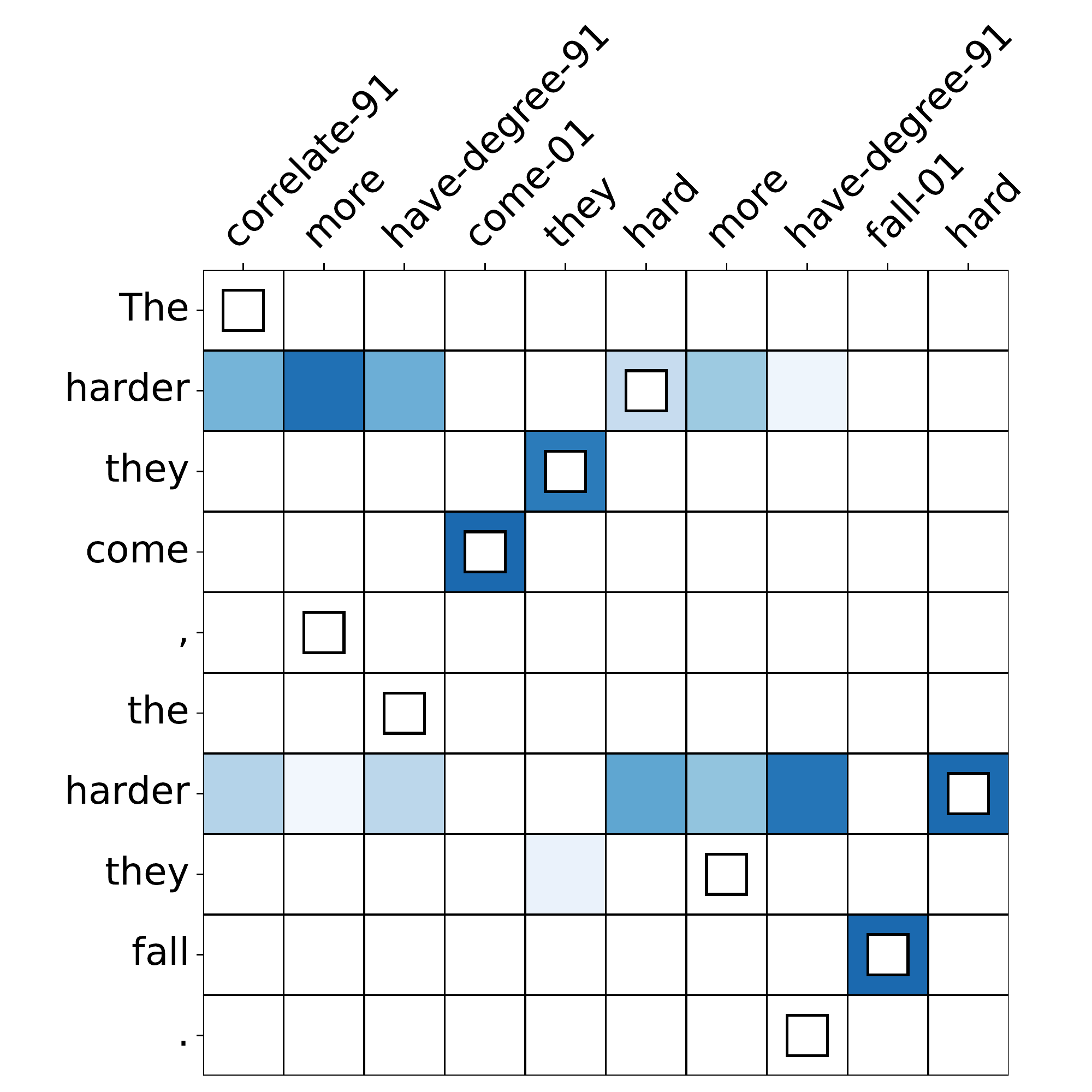}
\end{minipage}

\vspace{-2mm}
    \caption{(Left) An example of a sentence, its AMR graph, and the corresponding linearized AMR tree. The aligner decoder only incurs a loss for AMR nodes (tokens for nodes are in bold), although it represents the full history. (Right) A visualization of our alignment posterior (blue) and point estimate from the baseline (white box). The uncertainty corresponding to alignment ambiguity is helpful during sampling-based training.}
    \label{fig:amr_input_example} 
     \vspace{-3mm}
\end{figure*}
\vspace{-1mm}
\section{Inducing Alignments \label{neural_aligner}}
\vspace{-1mm}

Here, we describe our neural alignment model, which is essentially a variant of sequence-to-sequence models with hard attention \cite{yu-etal-2016-online,wu-etal-2018-hard,shankar2018,deng2018}. In contrast to SB-Align, our approach requires minimal pre-processing and does not have dependencies on many components or domain-specific rules.

The alignment model is trained separately from the AMR parser and optimizes the conditional likelihood of  \emph{nodes} in the linearized graph given the sentence.\footnote{This is because our oracle $O$ only needs node-word alignments to derive the oracle action sequence.}  The AMR graph is linearized by first converting the graph to a tree,\footnote{To convert the graph to a tree we only include the first incoming edge for each node.} and then linearizing the tree via a depth-first search, as in Figure~\ref{fig:amr_input_example}. Letting $v=v_1,\dots,v_{S}$ be the nodes in the linearized AMR graph, the log-likelihood is given by

\begin{align*}
   \log q(v \given w \param \phi) = \sum_{s=1}^{S} &\log q(v_s \given v_{<s}, w \param \phi),
\end{align*}
where we abuse notation and use $v_{<s}$ to indicate all the tokens (include brackets and edges) before $v_s$. That is, we incur losses only on the nodes $v_s$ but still \emph{represent} the entire history $v_{<s}$ for the prediction (see Figure~\ref{fig:amr_input_example}, left). The probability of each node is given by marginalizing over latent alignments $l_s$,
\begin{align*}
q(v_s \given v_{<s}, w \param \phi) =  \sum^{|w|}_{i=1}
                & q(l_s = i \given v_{<s} , w, \param \phi) \, \times \\
                & q(v_s \given l_s = i, v_{<s}, w \param \phi),
\end{align*}
where $l_s = i$ indicates that node $v_s$ is aligned to word $w_i$.

For parameterization, the sentence $w$ is encoded by a bi-directional LSTM. Each word is represented using a word embedding derived from a pretrained character-encoder from ELMo \cite{peters-etal-2018-deep}, which is frozen during training. On the decoder side, the linearized AMR tree history is represented by a uni-directional LSTM. The decoder shares word embeddings with the text encoder.  
The prior alignment probability $q(l_s = i \given v_{<s} , w \param \phi)$ is given by bilinear attention \cite{luong-etal-2015-effective},
\begin{align*}
    q(l_s = i \given v_{<s} , w \param \phi) &= 
    \frac{\exp(\alpha_{s, i})}{\sum^{|w|}_{j=1} \exp(\alpha_{s,j})}, \\
    \alpha_{s, i} &=
    h^{(v) \, \top}_{s} W h^{(w)}_{i},
\end{align*}

\noindent where $W$ is a learned matrix, $h^{(w)}_i$ is a concatenation of forward and backward LSTM vectors for the $i$-th word in the text encoder, and $h^{(v)}_{t}$ is the vector immediately before the $s$-th node in the graph decoder. 
The likelihood $q(v_s \given l_s = i, v_{<s}, w \param \phi)$ is formulated as a softmax layer with the relevant vectors concatenated as input,
\begin{align*}
    &q(v_s = y \given l_s = i, v_{<s}, w \param \phi) = \\
    &\qquad \mathrm{softmax}(U[h^{(y)}_{s}; h^{(w)}_{i}] + b)[y],
\end{align*}
\noindent where the softmax is over the node vocabulary and is indexed by the label $y$ belonging to the node $v_s$.

Once trained, we can tractably obtain the posterior distribution over each alignment $l_s$,
\begin{align*}
    &q(l_s = i \given w, v \param \phi) =  \\
    &\frac{q(l_s = i \given v_{<s} , w, \param \phi) \, q(v_s \given l_s = i, v_{<s}, w \param \phi)}{q(v_s \given v_{<s}, w \param \phi)},
\end{align*}
and the full posterior distribution over all alignments $l = l_1, \dots, l_S$ is given by 
\begin{align*}
    q(l \given w, g \param \phi) = \prod_{s=1}^S q(l_s  \given w, v \param \phi).
\end{align*}
\paragraph{Discussion}
Compared to the classic count-based alignment models, the neural parameterization makes it easy to utilize pretrained embeddings and also condition on the alignment and emission distribution on richer context. For example, our emission distribution $q(v_s \given l_s, v_{<s}, w \param \phi)$ can condition on the full target history $v_{<s}$ and the source context $w$, unlike count-based models which typically condition on just the aligned word $w_{l_s}$. In our ablation experiments described in (\S\ref{amr_parsing_analysis}) we find that the flexible modeling capabilities enabled by the use of neural networks are crucial for obtaining good alignment performance.

%% file: oraclesampling.tex
\section{Using Alignments \label{sec:use_align}}
\vspace{-1mm}
The neural aligner described above induces a posterior distribution over alignments, $q( l \given w, g \param \phi)$. We explore several approaches using this alignment distribution.
\vspace{-1mm}
\paragraph{MAP Alignment}

To use this alignment model in the most straightforward way, we decode the MAP alignment $\hat{l} = \argmax_l \, q( l \given w, g \param \phi)$ and train from the actions $\hat{a} = O(\hat{l}, w, g)$.

\paragraph{Posterior Regularization (PR)}
The action sequences derived from MAP alignments do not take into account the uncertainty associated with posterior alignments, which may not be ideal (Figure~\ref{fig:amr_input_example}, right). We propose to take this uncertainty into account and regularize the AMR parser's posterior to be close to the neural aligner's posterior at the distributional level. 

First, we note that the action oracle $\oracle$ is bijective as a function of $l$ (i.e., keeping $w$ and $g$ fixed), so the transition-based parser $p(a \given w \param \theta)$ induces a joint distribution over alignments and graphs,
\begin{align*}
    p(l, g \given w \param \theta ) \myeq  p(a = \oracle \given w \param \theta).
\end{align*}
This joint distribution further induces a marginal distribution over graphs,
\begin{align*}
    p(g \given w \param \theta) = \sum_{l} p(l, g \given w \param \theta),
\end{align*}
as well as a posterior distribution over alignments,
\begin{align*}
    p(l \given w, g\param \theta ) = \frac{p(l, g \given w \param \theta)}{p(g \given w \param \theta)}.
\end{align*}
A simple way to use the neural aligner's distribution, then, is via a \emph{posterior regularized} likelihood \cite{ganchev2010post},\footnotemark\footnotetext{Note that unlike in the original formulation, here we do not optimize over $q$ and instead keep it fixed. This is equivalent to the original formulation if we define the posterior regularization set $\mathcal{Q}$ to just consist of the distribution learned by the neural aligner, i.e.,  $\mathcal{Q} = \{q\}$.} 
\begin{align*}
    \mathcal{L}_\text{PR}(\theta) = \, &\log p(g \given w \param \theta) - \\ & \KL[{q}(l \given w, g \param \phi) \, \Vert \, p(l \given w, g \param \theta)].
\end{align*}
That is, we want to learn a parser that gives high likelihood to the gold graph $g$ given the sentence $w$ but at the same time has a posterior alignment distribution that is close to the neural aligner's posterior. Rearranging some terms, we then have
\begin{align*}
    \mathcal{L}_{\text{PR}}(\theta) = \, &\E_{q(l \given w, g \param \phi)}\left[\log p(l, g \given w \param \theta)\right] + \\ & \mathbb{H}[q(l \given w, g \param \phi)],
\end{align*}
and since the second term is a constant with respect to $\theta$, the gradient with respect to $\theta$ is given by,
\begin{align*}
    \nabla_\theta \mathcal{L}_{\text{PR}}(\theta) = \, &\E_{q(l \given w, g \param \phi)}\left[\nabla_\theta\log p(l, g \given w \param \theta)\right].
\end{align*}

Gradient-based optimization with Monte Carlo gradient estimators therefore results in an intuitive scheme where (1) we sample $K$ alignments $l^{(1)}, \dots, l^{(K)}$ from $q(l \given w, g \param \phi)$, (2) obtain the corresponding action sequences $a^{(1)}, \dots, a^{(K)}$ from the oracle, and (3) optimize the loss with the Monte Carlo gradient estimator $\frac{1}{K} \sum_{k=1}^K \nabla_\theta  \log p(a^{(k)} \given w, g \param \theta)$.

It is clear that the above generalizes the MAP alignment case. In particular, setting $q(l \given w, g) = \mathbbm{1}\{l = \hat{l}\}$ where $\hat{l}$ is the MAP alignment (or an alignment derived from the existing pipeline) recovers the existing baseline.

\paragraph{Importance Sampling (IS)}
The posterior regularized likelihood clearly lower bounds the log marginal likelihood $\mathcal{L}(\theta) = \log p(g \given w \param \theta),$\footnote{The log marginal likelihood is intractable to estimate directly due to the lack of any independence assumptions in the AMR parser, since in the AMR parser the alignment variable $l_s$ fully depends on $l_{<s}$.} and implicitly assumes that training against the lower bound results in a model that generalizes better than a model trained against the true log marginal likelihood.  In this section we instead take a variational perspective and use the neural aligner not as a regularizer, but as a \emph{surrogate posterior distribution} whose samples can be reweighted to reduce the gap between the $\mathcal{L}(\theta)$ and $\mathcal{L}_{\text{PR}}(\theta)$.

We first take the product of the neural aligner's posterior to create a joint posterior distribution,
\begin{align*}
  q(l^{(1)}, \dots l^{(K)} \param \phi) \myeq \prod_{k=1}^K q(l^{(k)} \given w, g \param \phi),
\end{align*}
where $K$ is the number of importance samples.
Then, \citet{burda2016importance} show that the following objective,
\begin{align*}
    \E_{q(l^{(1)}, \dots l^{(K)} \param \phi)}\left[\log \frac{1}{K}\sum_{k=1}^K \frac{p(l^{(k)}, g \given w \param \theta)}{q(l^{(k)}) \given w, g \param \phi) } \right],
\end{align*}
 motonotically converges to the log marginal likelihood $\log p(g \given w \param \theta)$ as $K \to \infty$. A single-sample\footnote{Note that a single sample from $q(l^{(1)}, \dots l^{(K)} \param \phi)$ is obtained by sampling from the neural aligner $K$ times.} Monte Carlo gradient estimator for the above is given by,
\begin{align*}
    \sum_{k=1}^K w^{(k)} \nabla_\theta \log p(a^{(k)} \given w, g \param \theta),
\end{align*}
where
\begin{align*}
    w^{(k)} = \frac{p(a^{(k)} \given w \param \theta) / q (l^{(k)} \given w, g \param \phi)}{\sum_{j=1}^{K} p(a^{(j)} \given w \param \theta) / q (l^{(j)} \given w, g \param \phi)}
\end{align*}
are the normalized importance weights \cite{ICML-2016-MnihR}. Thus, compared to the gradient estimator in the posterior regularized case which equally weights each sample, this importance-weighted objective approximates the true posterior $p(l \given w, g \param \theta)$ by first sampling from a fixed distribution $q(l \given w, g \param \phi)$ and then reweighting it accordingly.

\paragraph{Discussion} 
Despite sharing formulation with the variational autoencoder \cite{kingma2013auto} and the importance weighted autoencoder \cite{burda2016importance}, the approach proposed here differs in fundamental aspects. In contrast to the variational approaches we fix $q$ to the pretrained aligner posterior and do not optimize it further. Moreover, the lower bound $\mathcal{L}_{\text{PR}}(\theta)$ represents an inductive bias informed by a pretrained aligner, which can be more suited for early stages of training than even a tangent evidence lower bound (zero gap). This is because, for a tangent lower bound, $q$ in the Monte Carlo gradient estimate is equal to the true posterior over alignments for current model parameters. Since these parameters are poorly trained, it is easy for the aligner to provide a better alignment distribution for learning.

Posterior regularization seeks to transfer the neural aligner's strong inductive biases to the AMR parser, which has weaker inductive biases and thus may be potentially too flexible of a model. On the other hand, importance sampling ``trusts'' the AMR parser's inductive bias more, and uses the neural aligner as a surrogate distribution that is adapted to more closely approximate the AMR parser's intractable posterior. Thus, if the posterior regularized variant outperforms the importance sampling variant, it suggests that the StructBART is indeed too flexible of a model. While not considered in the present work, it may be interesting to explore a hybrid approach which first trains with posterior regularization and then switches to importance sampling.

%% file: experiments.tex
\section{Experimental Setup}

All code to run our experiments is available online\footnote{\href{https://github.com/IBM/transition-amr-parser}{https://github.com/IBM/transition-amr-parser}} with Apache License, 2.0. 

\subsection{Data, Preprocessing, and Evaluation}

\paragraph{Data} We evaluate our models on two datasets for AMR parsing in English. \textbf{AMR2.0} contains \textasciitilde39k sentences from multiple genres (LDC2017T10). \textbf{AMR3.0} is a superset of AMR2.0 sentences with approx. 20k new sentences (LDC2020T02), improved annotations with new frames, annotation corrections, and expanded annotation guidelines. Using AMR3.0 for evaluation allows us to measure how well our alignment procedure generalizes to new datasets --- AMR3.0 includes new sentences but also new genres such as text from LORELEI,\footnote{The LORELEI genre (low resource languages for emergent incidents) contains sentences from news articles, blogs, and forums \cite{strassel-tracey-2016-lorelei}. These sentences were specifically used in \citet{bevilacqua2021one} to measure parser out-of-domain generalization.} Aesop fables, and Wikipedia.

The primary evaluation of the aligner is extrinsically through AMR parsing, and we additionally evaluate alignments directly against ground truth annotations provided in \citet{blodgett-schneider-2021-probabilistic}---specifically, we look at the 130 sentences from the AMR2.0 train data (the ones most well suited for SB-Align), which we call the gold test set. Alignment annotations are not used during aligner training and only used for evaluation. 

\paragraph{Preprocessing} We align text tokens to AMR nodes. As the AMR sentences do not include de-facto tokenization, we split strings on space and punctuation using a few regex rules.

For AMR parsing we use the action set described in \S\ref{actionpointingoracle}. To accommodate the recurrent nature of the aligner, we linearize the AMR graph during aligner training. This conversion requires converting the graph into a tree and removing re-entrant edges, as described in \S\ref{neural_aligner}.

\paragraph{Evaluation} For AMR parsing we use Smatch \cite{cai-knight-2013-smatch}. For AMR alignment our goal is mainly to compare our new aligner with strong alignment baselines: SB-Align and LEAMR, a state-of-the-art alignment model \cite{blodgett-schneider-2021-probabilistic,austinThesis}.  However, our aligner predicts \textit{node-to-word} alignments, SB-Align predicts \textit{node-to-span} alignments, and the ground truth alignments  are \textit{subgraph-to-span}. To address this mismatch in granularity, we measure alignment performance using a permissive version of F1 after decomposing subgraph-to-span alignments into node-to-span alignments---a prediction is correct if it overlaps with the gold span. This permissiveness gives advantages the LEAMR and SB-Align baselines (which predict span-based alignments) as there is no precision-related penalty for predicting large spans.

\subsection{Models and Training}

\paragraph{Aligner} We use a bi-directional LSTM for the Text Encoder and uni-directional LSTM for the AMR Decoder. The input token embeddings are derived from a pretrained character encoder \cite{peters-etal-2018-deep} and frozen throughout training; these token embeddings are tied with the output softmax, allowing for alignment to tokens not seen during training. The alignment model is otherwise parameterized as described in \S\ref{neural_aligner}. We train for 200 epochs. Training is unsupervised, so we simply use the final checkpoint.\footnote{In our early experiments, we used SB-aligner's predictions as validation to find a reasonable range of  hyperparameters. Performance does not substantially deteriorate after 50 epochs, so this was not necessary or useful for early stopping. Early stopping based on perplexity performed similarly.} Additional training details for the aligner are in the Appendix.

\begin{table*}[t!]
\setlength\tabcolsep{4pt}
\begin{center}
\begin{tabular}{ l | c | c | c | c}
\toprule
Method & Beam Size & Silver Data  & AMR2.0 & AMR3.0  \\
\midrule
APT \cite{zhou-etal-2021-amr}$^{\mathcal{P}}$     & 10 & 70K          & 83.4                           & -  \\
TAMR \cite{xia-etal-2021-stacked-amr}$^{\mathcal{G}}$  &  8  & 1.8M         & 84.2                           & - \\
SPRING \cite{bevilacqua2021one}                   & 5  & 200K         & 84.3                            & 83.0 \\
StructBART-S \cite{zhou-etal-2021-structure}      & 10 & 90K          & -                               & 82.7 \small{$\pm0.1$} \\
StructBART-J \cite{zhou-etal-2021-structure}      & 10 & 90K          & 84.7 \small{$\pm$0.1}           & 82.6 \small{$\pm0.1$} \\
StructBART-J+MBSE \cite{lee2021maximum}           & 10 & 219K          & \textbf{85.7} \small{$\pm$0.0} & \textbf{84.1} \small{$\pm0.1$} \\
BARTAMR \cite{bai-etal-2022-graph}                &  5 & 200K          & 85.4                           & \textbf{84.2}\\ 

\midrule                                          
\midrule                                          
APT \cite{zhou-etal-2021-amr}$^{\mathcal{P}}$     & 10 &   -          & 82.8                           & 81.2  \\
SPRING \cite{bevilacqua2021one}                   & 5  &   -          & 83.8                           & \textbf{83.0} \\
SPRING \cite{bevilacqua2021one}$^{\mathcal{G}}$   & 5  &   -          & \textbf{84.5}                           & \textit{80.2} \\
StructBART-J \cite{zhou-etal-2021-structure}      & 10 &   -          & 84.2 \small{$\pm 0.1$}         & 82.0 \small{$\pm 0.0$} \\
StructBART-S \cite{zhou-etal-2021-structure}      & 10 &   -          & 84.0 \small{$\pm 0.1$}         & 82.3 \small{$\pm 0.0$} \\
\midrule                                                         
\midrule                                                         
StructBART-S (reproduced)                                      & 1 &   -  & 83.9 \small{$\pm 0.0$}         &  81.9 \small{$\pm 0.2$}\\
\hline
\: +neural-aligner (MAP)                                      & 1 &   -  & 84.0 \small{$\pm 0.1$} 	       & 82.5 \small{$\pm 0.1$} \\
\: +neural-aligner (MAP)                        & 10 &   -      & 84.1 \small{$\pm 0.0$} 	       & 82.7 \small{$\pm 0.1$} \\
\: +neural-aligner (PR, w/ 5 samples)  & 1 &   -  & \textbf{84.3} \small{$\pm 0.0$} 	       & \textbf{83.1} \small{$\pm 0.1$} \\
\: +neural-aligner (IS, w/ 5 samples)   & 1 &   -      & 84.2 \small{$\pm 0.1$} 	       & 82.8 \\
\bottomrule
\end{tabular}
\end{center}
\vspace{-2.5mm}
\caption{Results on parsing for AMR2.0 and 3.0 test sets. We report numbers when using single alignments (MAP), posterior regularization (PR), and importance sampling (IS). Also included are number of silver data training sentences used and beam size. PR and IS does not improve with beam search, and hence these numbers are omitted. $^{\mathcal{P}}$: Uses partial ensemble for decoder. $^{\mathcal{G}}$: Uses graph recategorization.}
\vspace{-4.5mm}
\label{tab:amr_parsing}
\end{table*}

\paragraph{AMR Parser} We use the StructBART model from \citet{zhou-etal-2021-structure} and the same hyperparameters: fine-tuning for $100$ epochs (AMR2.0) or $120$ (AMR3.0), and using Smatch on the validation set for early stopping. We did not tune the hyperparameters of the AMR parser at all as we wanted to see how well the neural aligner performed as a ``plug-in'' to an existing system. Additional implementation details for parsing are in the Appendix. For posterior regularization (PR) and importance sampling (IS) variants we use $K=5$ samples to obtain the gradient estimators.

\vspace{-1mm}
\section{Results and Analysis}
\vspace{-1mm}
\label{amr_parsing_analysis}

The full results are shown in Table~\ref{tab:amr_parsing}, where we find that our approach can learn state-of-the-art AMR parses for gold-only training and without requiring beam search. We now interpret the main results in more detail.

\vspace{-1mm}
\paragraph{Pipeline Generalization} SB-Align was developed prior to the AMR3.0 release, and because it incorporates a complex pipeline with domain-specific rules, one could argue it is specialized for prior datasets like AMR2.0. In Table \ref{tab:amr_parsing} our aligner yields relatively stronger StructBART improvements for AMR3.0 than AMR2.0. This result and the relatively little manual configuration our aligner requires (e.g., no rules, lemmatization, etc.) suggest our alignment approach generalizes better to different training corpora and that prior peformance of StructBART on AMR3.0 may have been affected by a lack of generalization.

Graph re-categorization \cite{zhang-etal-2019-amr,zhang-etal-2019-broad} is a commonly used technique in AMR parsing where groups of nodes are collapsed during training and test time, but expanded during evaluation. Results from \cite{bevilacqua2021one} show re-categorization may be harmful, but our results suggest a different perspective---re-categorization is partially a function of alignment-like heuristics and the lower re-categorization results of SPRING in AMR3.0 reinforce our findings that alignments based on heuristic are difficult to generalize.

\vspace{-1mm}
\paragraph{Alignment Uncertainty vs. Data Augmentation} We improve parsing performance by sampling 5 alignments per sentence in batch (see Table \ref{tab:amr_parsing}). One can argue that our approach simply exposes the model to more data, but we found that training on one-best alignments for longer did not improve results.  When looking at our sampling results compared with previous versions of StructBART trained on silver data, we see that our approach even outperforms the benefit of the simpler versions of data augmentation, such as simple self-learning. This suggests that there is possible further improvement by combining both techniques, which we leave for future work.

\paragraph{Posterior regularization vs. Importance sampling}
Training with posterior regularization  or importance sampling uses the same number of samples, but in different ways. In posterior regularization, the samples are used to better approximate the posterior regularized objective, which in turn regularizes the AMR parser's posterior more effectively by reducing the gradient estimator variance. In importance sampling, the samples are used to better approximate the AMR parser's intractable log marginal likelihood. We find that importance sampling fails to improve upon posterior regularization for both AMR2.0 and AMR3.0, which indicates that strong inductive biases associated the constrained aligner is a useful training signal for the flexible AMR parser.

\begin{table}[t]
\setlength\tabcolsep{4pt}
\begin{center}
\begin{tabular}{ l | c }
\toprule
 & AMR2.0 \\
\midrule
LEAMR \small{\cite{blodgett-schneider-2021-probabilistic}} & 97.4 \\
SB-Align \small{\cite{zhou-etal-2021-structure}}             & 89.2  \\
Neural Aligner (ours)  & 96.5 \\
\midrule
IBM Model 1   & 77.2  \\
Neural Aligner w/o pretrained emb.    & 79.8 \\
Neural Aligner w/ pretrained emb.  & 96.5 \\
\bottomrule
\end{tabular}
\end{center}
\vspace{-2mm}
\caption{(Top) Alignments results using ground truth data from \citet{blodgett-schneider-2021-probabilistic}, where we use the 130 sentences from the gold test set that are from the AMR2.0 train data. (Bottom) Alignment ablation results against the same test set using different alignment model variants. Note that as the ground truth alignment is at the span level, we report a permissive variant of F1 where a prediction is considered correct if it partially overlaps with the ground truth span. This advantages the LEAMR and SB-Align baselines as these can align to spans, whereas our aligners only align to words.}
\label{tab:amr_alignment_results}
\end{table}

\paragraph{Comparing against alignment baselines}

Our neural alignment method is preferred over SB-Align for two primary reasons: it is relatively easy to use (makes use of word embeddings, depends on less preprocessing, does not require domain-specific rules, etc.) and empirically improves performance (see Table \ref{tab:amr_parsing}). Nonetheless, we conduct an intrinsic evaluation to assess the quality of the predicted alignments---it is desirable that our aligner actually provides accurate alignments.

To verify that improved parsing is due to better alignment, we compare against two strong alignment baselines (LEAMR and SB-Align) on an evaluation set of gold manually annotated alignments. In general, there are only a few hundred such annotations available, yet we aim to use these alignments on 10s or 100s of thousands sentences for AMR parsing. For this reason all the aligners are trained unsupervised with respect to alignment. The results in Table \ref{tab:amr_alignment_results} (top) show our aligner is substantially better than SB-Align and nearly on-par with LEAMR, the current state of the art.

\paragraph{Aligner Parameterization}

We train the classic count-based IBM Model 1 \cite{brown1993mathematics} using expectation maximization.  We next train our neural aligner without pretrained character-aware embeddings. Our neural aligner is different from the classic IBM model in that (1) it learns the prior alignment distribution, (2) the emission model conditions on the entire sentence $w$ and the target history $v_{<s}$. Finally, adding pretrained embeddings to this model recovers our full model. The results in Table \ref{tab:amr_alignment_results} (bottom) indicate that both flexibility and token representation are required to outperform IBM Model 1. Training with word vectors learned from scratch only provides a small benefit, and the best performance is from using pretrained character embeddings, which yields nearly 20 point improvement in our permissive F1 metric.

%% file: relatedwork.tex
\section{Related Work}

\paragraph{Oracles for parsing} Dynamic oracles in syntactic parsing \cite[\textit{inter-alia}]{goldberg-nivre-2012-dynamic,ballesteros-etal-2016-training} enable oracles to recover from imperfect sequences. Oracles with random exploration in transition-based AMR parsers have been previously explored using imitation-learning \cite{goodman-etal-2016-noise}, reinforcement learning \cite{naseem-etal-2019-rewarding} and oracle mining \cite{lee-etal-2020-pushing}. In addition to this, \citet{liu-etal-2018-amr} produces multiple alignments via their rule-based system and selects the best based on parser performance. Compared to our proposed posterior regularization training, dynamic oracle works exploit specific linguistic properties of syntactic trees not directly transferable to AMR graphs. Prior work on random exploration and AMR selects oracles based on the Smatch. Our work uses the action space and oracle from StructBART \cite{zhou-etal-2021-structure,zhou-etal-2021-amr}, which requires every node to be aligned to a word so that the AMR graph is fully recoverable. We expose the parser to uncertainty by sampling alignment, which does not require computing the Smatch metric. Rather, the aligner is trained separately from the parser using pairs of sentences and their respective AMR graphs.

\paragraph{Neural Alignments} The neural aligner we use is closely related to sequence-to-sequence models with hard (i.e., latent variable) attention \cite[][\textit{inter alia}]{Xu2015,Ba2015,wu-etal-2018-hard,shankar2018,deng2018,shankar2019} and other works on marginalizing over monotonic alignments \cite{yu2016online,pmlr-v70-raffel17a,wu-cotterell-2019-exact}.  In these works, the main goal is to obtain better sequence-to-sequence models, and the actual posterior distribution over alignments is a byproduct (rather than the goal) of learning. However, there are also works that utilize contemporary neural parameterizations and explicitly target alignments as the primary goal \cite{zenkel2020align,ho2020align,chen2021align}. Prior work on integrating pretrained aligners with sequence-to-sequence models has generally used the alignments to supervise the intermediate soft attention layers \cite{liu-etal-2016-neural,cohn-etal-2016-incorporating,yin-etal-2021-compositional}, in contrast to the present work which formally treats alignments as latent variables.

\paragraph{Alignments in AMR Parsing} There is extensive work aligning AMR nodes to sentences. The StructBART model \cite{zhou-etal-2021-structure} considered here, previously made use of well established rule-based and statistical AMR aligners \cite{flanigan-etal-2014-discriminative,pourdamghani-etal-2014-aligning} trained with expectation maximization and additional heuristics \cite{naseem-etal-2019-rewarding,emnlp2020stacktransformer} that we call SB-Align. Our aligner compares favorably against SB-Align in alignment metrics and downstream parsing performance. The predicted alignments we use come close to \citet{blodgett-schneider-2021-probabilistic} measured on gold alignment annotations, despite our method leveraging less domain-specific components. \citet{lyu-titov-2018-amr} and \citet{lyu-etal-2021-differentiable} incorporate differentiable relaxations of alignments for AMR parsing.

\paragraph{Posterior Regularization for Parsing and Generation} \citet{Li_Cheng_Liu_Keller_2019} apply variational inference and posterior regularization for unsupervised dependency parsing using their transition-based system. Their approach predates large pretrained language models for which the use of structure may play a different role. \citet{li-rush-2020-posterior} use posterior regularization to incorporate weakly supervised alignment constraints for data-to-text generation, also without pretrained neural representations in mind.